%% file: main.tex
\documentclass[10pt,twocolumn,letterpaper]{article}

\usepackage[pagenumbers]{wacv} 

\input{preamble}

\usepackage[pagebackref,breaklinks,colorlinks]{hyperref}

\usepackage[shortlabels]{enumitem}

\usepackage[capitalize]{cleveref}
\crefname{section}{Sec.}{Secs.}
\Crefname{section}{Section}{Sections}
\Crefname{table}{Table}{Tables}
\crefname{table}{Tab.}{Tabs.}

\def\wacvPaperID{2} 
\def\confName{WACV}
\def\confYear{2024}

\def\sname{EarlyBird\xspace}

\begin{document}

\title{EarlyBird: Early-Fusion for Multi-View Tracking in the Bird’s Eye View}

\author{
Torben Teepe\thanks{Correspondence to \texttt{t.teepe@tum.de}} \qquad Philipp Wolters \qquad Johannes Gilg \qquad Fabian Herzog \qquad Gerhard Rigoll\\[0.1cm]
Technical University of Munich
}
\maketitle

\input{sec/0_abstract}
\glsresetall

\input{sec/1_intro}

\input{sec/2_related}
\input{sec/3_main}
\input{sec/4_experiments}

\input{sec/5_conclusion}

{\small
\bibliographystyle{ieee_fullname}
\bibliography{main}
}

\end{document}

%% file: preamble.tex
\usepackage{amsmath}
\usepackage{amssymb}
\usepackage{bm}
\usepackage{booktabs}
\usepackage{ulem}
\usepackage{adjustbox}
\usepackage[dvipsnames]{xcolor}
\usepackage{siunitx}
\usepackage{nicefrac}
\usepackage{multirow}

\newcommand{\nparagraph}[1]{\noindent\textbf{#1.  }}

\usepackage[acronym]{glossaries}
\newacronym{mtmc}{MTMC}{Multi-Target Multi-Camera}
\newacronym{bev}{BEV}{Bird’s Eye View}
\newacronym{moda}{MODA}{Multiple Object Detection Accuracy}
\newacronym{modp}{MODP}{Multiple Object Detection Precision}
\newacronym{iou}{IoU}{Intersection over Union}
\newacronym{fov}{FOV}{Field of View}
\newacronym{mot}{MOT}{Multiple Object Tracking}
\newacronym{mota}{MOTA}{Multiple Object Tracking Accuracy}
\newacronym{motp}{MOTP}{Multiple Object Tracking Precision}
\newacronym{reid}{re-ID}{Re-Identification}
\newacronym{pom}{POM}{probabilistic occupancy map}
\newacronym{nms}{NMS}{non-maximum suppression}

\usepackage{xspace}

%% file: sec/0_abstract.tex
\begin{abstract}
    Multi-view aggregation promises to overcome the occlusion and missed detection challenge in multi-object detection and tracking.
    Recent approaches in multi-view detection and 3D object detection made a huge performance leap by projecting all views to the ground plane and performing the detection in the \gls{bev}. 
    In this paper, we investigate if tracking in the \gls{bev} can also bring the next performance breakthrough in \gls{mtmc} tracking.
    Most current approaches in multi-view tracking perform the detection and tracking task in each view and use graph-based approaches to perform the association of the pedestrian across each view. This spatial association is already solved by detecting each pedestrian once in the \gls{bev}, leaving only the problem of temporal association.
    For the temporal association, we show how to learn strong \gls{reid} features for each detection.
    The results show that early-fusion in the \gls{bev} achieves high accuracy for both detection and tracking.
    \sname outperforms the state-of-the-art methods and improves the current state-of-the-art on Wildtrack by $+4.6$ MOTA and $+5.6$ IDF1.
    \iftoggle{wacvfinal}{%
    \href{https://github.com/tteepe/EarlyBird}{https://github.com/tteepe/EarlyBird}
    }{\href{https://anonymous.4open.science/r/EarlyBird-3641/}{https://anonymous.4open.science/r/EarlyBird-3641/}.}
\end{abstract}

%% file: sec/1_intro.tex
\section{Introduction}\label{sec:intro}

\begin{figure}[t]
  \centering
  \vskip0.5em
   \includegraphics[width=\linewidth]{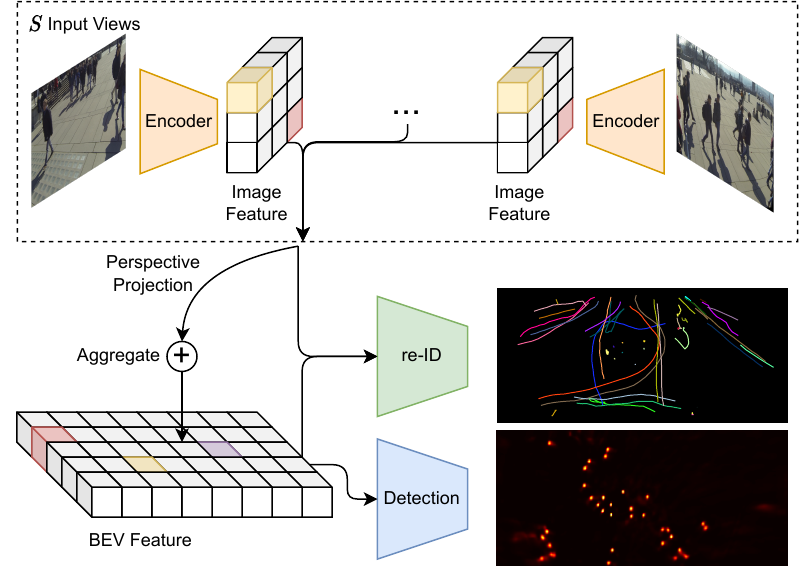}
   \vskip0.25em
   \caption{Overview of our approach. All input images are encoded and then perspectively projected to the ground plane. The aggregation reduces the BEV feature where we detect pedestrians and predict a \acrshort{reid} feature for tracking.}
   \label{fig:teaser}
\end{figure}

Detection and tracking of pedestrians has been an essential problem with numerous applications in video surveillance, autonomous vehicles, and sports analysis. Despite the progress on monocular \gls{mot} occlusion remains one of the biggest challenges in this research field. Occlusion causes detections to get lost and tracks to get fragmented, thus limiting the detection and tracking quality. However, practical situations like sports analysis require detections in highly cluttered or crowded scenes. Multiple cameras with an overlapping field of view might be available for these cases. Observing a scene from multiple views can help overcome these occlusions since objects hidden in one camera can be visible in another. The challenge then is to aggregate information from multiple camera views. 
In early approaches multi-view detection was solved with late fusion methods \cite{xu2016multi}: First, pedestrians are detected in a single view, then this detection is projected to the 3D space or mostly the ground plane where it is associated with the projections of the other views. More recent approaches \cite{hou2020multiview, hou2021multiview} utilize an early-fusion strategy that first projects a representation of all views to the common ground plane or \textit{\glsentrylong{bev}} and then perform the detection. These early-fusion detectors \cite{hou2020multiview, hou2021multiview} increased the detection quality significantly compared to the previous late-fusion approaches. Late-fusion approaches commonly have the advantage that they require less hardware because the processing can be performed independently, and the information projected to 3D is more sparse than the full images. Early-fusion approaches have the advantage that they can be trained end-to-end, while late-fusion usually optimizes the detection and the multi-view association separately.
A challenge for the detection in the \gls{bev}-space has been the distortion created by the perspective transformation. Several approaches\cite{hou2021multiview, lee2023multi, song2021stacked} tried to overcome this problem. We build our approach on \cite{hou2020multiview} but add a \gls{bev}-Decoder that is based on a ResNet-18 and gives the decoded features a larger receptive field, allowing the model to aggregate information from the distortion \textit{shadows} to the actual location. We mainly focus on the tracking task, but our model also achieves competitive results in the detection task.

While early-fusion has been shown to be the stronger approach for detection, tracking in multi-view is still performed with the late-fusion approach \cite{hofmann2013hypergraphs, cheng2023rest}: first 2D detections are acquired. Secondly, detections of each timestep are associated, and finally, the detections are associated across timesteps. Other approaches \cite{synthehicle2023herzog, nguyen2022lmgp} switch the order and first associate within one view and later match these tracks across the views. Regardless of the ordering, any stage in this tracking pipeline suffers from inaccuracies introduced by the prior stage, i.e., missed 2D detection later needs to be compensated in the association stage. Our approach combines the first two steps and directly performs the detections in the \gls{bev} building upon the latest multi-view detectors \cite{hou2021multiview}. For tracking, we adopt the idea introduced by FairMOT~\cite{zhang2021fairmot} and simultaneously learn a \gls{reid} feature for each detection in the \gls{bev}-space. This approach allows us to skip the first step of spatial association since our learned detector already solves this problem. The associate in the temporal domain is first performed with appearance-based \gls{reid} features and secondly with a Kalman filter \cite{kalman1960new} as a motion-based model. We call this architecture \sname. It is an online, end-to-end, trainable tracking architecture that improves the state-of-the-art in tracking by a large margin.

Our contributions are the following:
\begingroup
\renewcommand\labelenumi{\theenumi)}
\begin{enumerate}[nosep]
\item We introduce early-fusion tracking in the \glsentrylong{bev} with a simple but strong \gls{reid} association strategy.

\item We introduced a more robust decoder architecture for the \gls{bev} features that improve our tracking results and detections. 

\item  In our experiments, we qualitatively and quantitatively verify the effectiveness of our method against recent relevant methods and improve the state-of-the-art in tracking on \textit{Wildtrack} by a $+4.6$ MOTA and $+5.6$ IDF1.
\end{enumerate}
\endgroup

%% file: sec/2_related.tex
\section{Related Work}\label{sec:related}

\nparagraph{Multi-View Object Detection}
Using a multiple-camera setup is a prevalent solution to address the difficulties of pedestrian detection with heavy occlusions. Such a setup utilizes synchronized and calibrated cameras observing the same area from different perspectives. The multi-view detection system then integrates these images, all of which have overlapping fields of view, to perform pedestrian detection. Probabilistic modeling of objects \cite{coates2010multi, sankaranarayanan2008object} was the primary focus before the advancements brought by deep learning. Techniques such as mean-field inference \cite{baque2017deep, fleuret2007multicamera} and conditional random field (CRF) \cite{baque2017deep, roig2011conditional} were commonly employed for the aggregation of information from multiple views. However, these techniques often necessitated additional computations or specific designs not inherent in deep learning models. MVDet \cite{hou2020multiview} proposed a convolution-based, end-to-end trainable method that projects encoded image features from each view to the common ground plane, yielding significant improvements and making it the base architecture for all following approaches, including ours. Instead of projecting only the sparse detection from each view to the ground plane, \cite{hou2020multiview} first applies an encoder to the input image and projects all features to the ground plane with perspective transformation. The perspective transformation, which projects image features that depict areas over the ground plane of the actual location in 3D, causes distortions in the ground plane resembling a shadow of the actual object \cite{hou2021multiview}. Other approaches \cite{hou2021multiview, lee2023multi, song2021stacked} try to overcome these shortcomings of the perspective transformation: \cite{hou2021multiview} uses projection-aware transformers with deformable attention in the \gls{bev}-space to aggregate those \textit{shadows} back to the original location. \cite{lee2023multi} uses regions of interest from the 2D detections and separately projects those to the estimated foot location on the ground plane. Another approach \cite{song2021stacked} aims to overcome the shortcomings of perspective transformation by using multiple stack homographies at different heights to approximate a complete 3D projection.
Instead of focusing on the model side, \cite{qiu20223d} tried to improve detection on the data side. This approach added additional occlusions with 3D cylindrical objects. This data augmentation makes it harder for the approach to always rely on multiple cameras and thus helps to avoid overfitting.

Our approach builds upon MVDet \cite{hou2020multiview} because it is a solid and straightforward baseline for early-fusion multi-view object detection that we can extend with our tracking approach. 

\begin{figure*}[t]
  \centering
   \includegraphics[width=\linewidth]{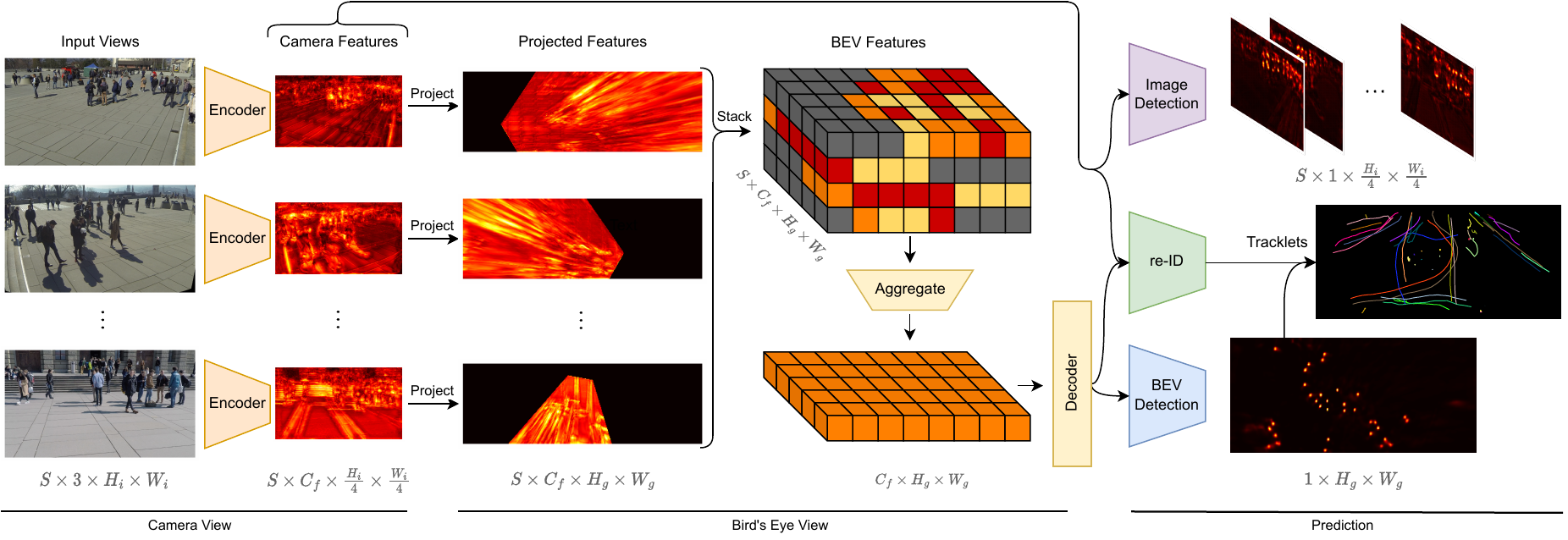}

   \caption{Overview of our approach. The input view are encoded and the resulting camera features are projected to the ground plane. The projected features are then stacked and aggregated to yield the BEV feature. For the image features the box centers are predicted to guide the occupancy detection in the BEV. Additionally we train a \gls{reid} feature that is guided both by the camera features as well as the BEV features. The detections and their corresponding \gls{reid} features are then used to associate the detections into tracklets.}
   \label{fig:overview}
\end{figure*}

\nparagraph{Multi-Target Multi-Camera Tracking}
There is much literature on single-camera tracking, and we will discuss one-shot trackers later, but in this section, we discuss the relevant works in \gls{mtmc} tracking. Most of \gls{mtmc} trackers assume an overlapping \gls{fov} between the cameras. Fleuret \etal \cite{fleuret2007multicamera} use the overlapping \gls{fov} to model targets into a \gls{pom} and combine occupancy probabilities with color and motion attributes in the tracking process. As an improvement \cite{berclaz2011multiple}, formulate tracking in \gls{pom}s as an integer programming problem, and compute the optimal solution by using the k-shortest paths (KSP) algorithm.
The problem of \gls{mtmc} tracking can also be seen as a graph problem. Hypergraphs \cite{hofmann2013hypergraphs} or multi-commodity network flows \cite{shitrit2013multi, leal2012branch} are used to model the correspondences across the views and then solved with min-cost \cite{hofmann2013hypergraphs, shitrit2013multi} or with branch-and-price algorithms \cite{leal2012branch}.

In recent years, a two-step approach has become popular \cite{synthehicle2023herzog}: first generating local tracklets of all the targets within each camera, later matching local tracklets that belong to the same target across all the cameras. For the first step, the generation of local tracklets within a single camera is referred to as the single camera \gls{mot}, which has been studied intensively \cite{bergmann2019tracking, zhang2021fairmot, zhou2020tracking, wang2020towards, feichtenhofer2017detect, Wojke2017simple, chen2018real}. Due to the impressive progress of object detection techniques, tracking-by-detection \cite{feichtenhofer2017detect, seidenschwarz2023simple,bergmann2019tracking, zhou2020tracking, Wojke2017simple} has become the mainstream approach for multi-target tracking in recent years. For the second step, various cross-view data association methods have been proposed to match local tracklets across different cameras. Some works \cite{hu2006principal, eshel2008homography} use the properties of the epipolar geometry to find correspondences based on location on the ground plane. In addition to ground plane locations \cite{xu2016multi} adds appearance features as cues for the association.
The current state-of-the-art models \cite{nguyen2022lmgp, cheng2023rest} flip the first two steps: the 2D detections are first projected to the 3D ground plane, and a graph is constructed with \gls{reid} node features. The nodes are then either first assigned spatially and temporally \cite{cheng2023rest} or both assignments happen in the same step \cite{nguyen2022lmgp} using graph neural networks for link prediction. While all current approaches \cite{cheng2023rest, seidenschwarz2023simple,bergmann2019tracking, zhang2021fairmot, zhou2020tracking} evaluate on detection results to also account for detection inaccuracies, LMGP \cite{nguyen2022lmgp} evaluates on groundtruth bounding boxes and thus can not be compared to any recent works.
Our approach differs from all previous work and is more comparable to one-shot trackers covered in the next section. Our approach shares the idea with latest approaches \cite{nguyen2022lmgp, cheng2023rest} to first associate spatially in our detector and then associate on the ground plane.  

\nparagraph{One-Shot Tracking}
A special case of single-view Multi-Object Trackers is one-shot trackers. These trackers perform the detection and tracking in one step, thus reducing inference time. They usually have a lower performance compared to two-step trackers. The features predicted can ether be \gls{reid} feature \cite{zhang2021fairmot, Voigtlaender19mots, wang2020towards} or motion cues \cite{zhou2020tracking, bergmann2019tracking, feichtenhofer2017detect}. The first example for a \gls{reid}-based approach is Track-RCNN \cite{Voigtlaender19mots} that adds a \gls{reid} feature extraction on top of Mask R-CNN \cite{he2017mask} and regresses a bounding box and a re-ID feature for each proposal. Similarly, JDE \cite{wang2020towards} is build upon YOLOv3\cite{redmon2018yolov3}, and FairMOT is build upon CenterNet \cite{zhou2019objects}. The advantage of FairMOT compared to the others is that it is anchor-free, meaning detections are not based on bounding boxes but on a single detection point, leading to better separation of the \gls{reid} features.
D\&T was proposed as a motion-based tracker in  \cite{feichtenhofer2017detect}, which takes input from adjacent frames and predicts inter-frame offsets between bounding boxes. Tracktor \cite{bergmann2019tracking} directly exploits the bounding box regression head to propagate identities of region proposals and thus removes box association. Unlike other methods, CenterTrack \cite{zhou2020tracking} predicts the object center offset on a triplet input: current frame, last frame, and the heatmap of last frame detection. The previous heatmap allows this method to match objects anywhere, even if the boxes overlap.
However, motion-based methods only associate objects in adjacent frames without re-initializing lost tracks and thus have difficulty handling occlusions.

In our approach, we thus bring the concept of joint detection and \gls{reid} extraction from FairMOT \cite{zhang2021fairmot} to \gls{mtmc} tracking. While training \gls{reid} features for images is well-understood task \cite{herzog2021lightweight, zhang2021fairmot, Voigtlaender19mots, wang2020towards}, projecting strong \gls{reid} features to the \gls{bev} is what we will investigate in this work.

%% file: sec/3_main.tex
\section{EarlyBird}\label{sec:main}

We provide a comprehensive overview of \sname in \cref{fig:overview}. It starts with the input images that are augmented and fed to the encoder network to yield our image features. The image features have the size of the input images downsampled by 4. The image features from all cameras are subsequently projected to the ground plane and stacked into the \gls{bev} space. In the following step the \gls{bev} space is then reduced in the vertical dimension. The \gls{bev} features are finally fed through a decoder network. Both image features and \gls{bev} features have separate heads for center and offset detection but share a head for \gls{reid} prediction. 


\subsection{Encoder}
Our approach assumes synchronized RGB input images from $S$ cameras with an input size of $3 \times H_i \times W_i$. We encode the features of the images with ResNet or Swin Transformer networks using three blocks of the network, with each block downsampling the input by $2$. Our goal is to only downscale the images by the factor of $4$, and thus we upsample and concatenate the output features of each layer until we get an output of $C_f \times H_f \times W_f$ with $H_f = H_i / 4$, $W_f = W_i / 4$ and $C_f = 128$.

\subsection{Projection}
The projection is the central part of this approach as it gives a parameter-free link between the image view and the \gls{bev}-view. Following \cite{hou2020multiview}, we use perspective projection to project the image features to the ground plane. Using the pinhole camera model \cite{hartley2003multiple}, translation between 3D locations $\left(x,y,z\right)$ and 2D image pixel coordinates $\left(u,v\right)$ are calculated with:
\vspace{-1em}\begin{equation}
\label{eq:perspective_4x3}
\resizebox{0.9\linewidth}{!}{%
    $s\begin{pmatrix}u \\ v \\ 1\end{pmatrix} = %
    \bm{K} \left[ \bm{R}|\bm{t}\right] \begin{pmatrix} x \\ y \\ z \\ 1\end{pmatrix}\\ = \begin{bmatrix} p_{11} & p_{12} & p_{13} & p_{14} \\ p_{21} & p_{22} & p_{23} & p_{24} \\ p_{31} & p_{32} & p_{33} & p_{34} \end{bmatrix} \begin{pmatrix} x \\ y \\ z \\ 1\end{pmatrix},$
}\end{equation}
where $s$ is a real-valued scaling factor, $\bm{P} = \bm{K} \left[ \bm{R}|\bm{t}\right]$ is a $3\times4$ perspective transformation matrix, $\bm{K}$ are intrinsic camera matrix and, $\left[\bm{R}|\bm{t}\right]$ is the $3\times4$ extrinsic parameter matrix. \cref{eq:perspective_4x3} describes the ray corresponding to each pixel $\left(u,v\right)$ in the 3D world. In our approach, we choose to project all pixels to the ground plane $z=0$, then the projection can be simplified to:%
\vspace{-0.5em}\begin{equation}
\label{eq:perspective_3x3}
\resizebox{0.78\linewidth}{!}{$
    s\begin{pmatrix}u \\ v \\ 1\end{pmatrix} = \bm{P_0} \begin{pmatrix} x \\ y \\ 1\end{pmatrix} = \begin{bmatrix} p_{11} & p_{12} & p_{14} \\ p_{21} & p_{22} & p_{24} \\ p_{31} & p_{32} & p_{34} \end{bmatrix} \begin{pmatrix} x \\ y \\ 1\end{pmatrix},
$}\end{equation}
where $\bm{P_0}$ denotes the $3\times3$ perspective transformation matrix without the third column from $\bm{P}$. We apply \cref{eq:perspective_3x3} to project the features from all $S$ cameras, with their projection $\bm{P_0^{(s)}}$, to the ground plane grid of a predefined size $\left[H_g,W_g\right]$. The size of the ground plane grid depends on the size of the observed and annotated area. Each grid position represents an area of $\qty{10}{\cm} \times \qty{10}{\cm}$, downsampling the annotation grid further by $4$ due to memory concerns. All stacked feature maps with $C$-channels from $S$ cameras give us \gls{bev} feature of size $S\times C_f \times H_g \times W_g$. 

\subsection{Aggregation \& Decoder}
The goal of the aggregation stage is to combine the features from all $S$ cameras into a single feature, i.e., reduce the $S$-dimension of the \gls{bev} feature map. We concatenate all feature maps along the channel dimension, as in $S\times C_f \times H_g \times W_g \rightarrow (S \cdot C_f) \times H_g \times W_g$, yielding a high-
dimensional \gls{bev} feature map. With two 2D convolutions, we reduce this high-dimensional \gls{bev} feature to our desired channel size of $C_g = 128$.

After the aggregation, we feed the \gls{bev} feature into a ResNet-18 decoder. The goal of the decoder is to introduce a large receptive field of the ground plane. The distortion introduced by the perspective projections causes pedestrian features to spread out from their actual location on the ground plane. Other approaches \cite{hou2021multiview, song2021stacked, qiu20223d, lee2023multi} identified this distortion as harmful to the detection accuracy and all proposed complex solutions, like deformable transformers \cite{hou2021multiview} or ROI projection \cite{lee2023multi}. Our decoder offers a simple solution to aggregate location and identification features on the ground plane.

In each layer of the ResNet, the \gls{bev} feature is downsampled by $2$. We then use a pyramid network architecture to upsample the output of each layer to the size of the previous larger output. Then, both features are concatenated in the channel dimension, and a 2D convolution is applied. The feature pyramid yields a decoded output with the same shape as the input of $C_g \times H_g \times W_g$ but a much higher receptive field for each grid location. 

\input{tab/detection}

\subsection{Heads \& Losses}
To get the final prediction of the \gls{pom}, we use prediction heads on our \gls{bev} feature map. The detection architecture follows CenterNet \cite{zhou2019objects}, and we add a head for center detection that reduces the feature to $1 \times H_g \times W_g$ to yield a heatmap or \gls{pom} on the ground plane. We add another head for offset prediction that helps predict the location more accurately as it mitigates the quantization error from the ground grid. The offset has an $(x,y)$ component and has the shape  $1 \times H_g \times W_g$. Each head is implemented by applying a $3 \times 3$ convolution (with $C_g = 128$ channels), followed by and activation layer and a $1 \times 1$ convolution to the final target size. The center head is trained with Focal Loss, and the offset head is trained with L1 Loss.

We also add detection heads for image features that predict the center of the 2D bounding boxes and estimated foot location at the bottom-center of the bounding box, helping the image features to have higher activations at the location of each pedestrian.
Following FairMOT \cite{zhang2021fairmot}, we add an uncertainty term to automatically balance the single-task losses before summing them up.

\nparagraph{Re-Identification}
The \gls{reid} head aims to generate features that can distinguish individual pedestrians. Ideally, affinity among different pedestrians should be smaller than between the same pedestrian. To archive this, we learn \gls{reid} features through a classification task and as a metric learning task. First, we apply a head that yields the \gls{reid} feature on the ground plane $C_{id, g} \times H_g \times W_g$ with $C_{id} = 64$ and also of the image features $C_{id, f} \times H_f \times W_f$. Afterwards, we extract the feature at the location of the center detection in both planes. We create a class identity distribution with a linear layer that we train with Cross Entropy Loss to the ground truth class identity. As discussed earlier, the perspective transformation introduces strong distortion on the ground plane. Thus, we supervise the \gls{reid} features from the image view. In addition to the Cross-Entropy loss, we apply SupCon Loss \cite{khosla2020supervised}, which pulls features belonging to the same class identity together while simultaneously pushing apart features of samples from different classes.

\input{tab/tracking}

\subsection{Inference}
At inference time, we take the \gls{pom} predicted by the \gls{bev} center head and perform \gls{nms} by a simple $3 \times 3$ max pooling operation as in \cite{zhou2020tracking}. We then only extract the detections over a certain threshold of $0.4$. We also extract the identity embeddings at the estimated pedestrian centers. In the next section, we discuss how we associate the detected boxes over time using the re-ID features.

\nparagraph{Online Association}
We adopt the hierarchical online data association approach described by MOTDT \cite{chen2018real}, but instead of boxes, we only track the pedestrians centers seen from the \textit{\glsentrylong{bev}}. Our first step involves initializing a set of tracklets based on the centers detected in the initial timestep. As each subsequent timestep is processed, we connect the centers detected to the existing tracklets using a two-stage matching strategy.

In the first stage, we use a combination of the Kalman Filter \cite{kalman1960new}, and re-ID features to achieve initial tracking results. Specifically, we use the Kalman Filter to anticipate tracklet locations in the next frame and calculate the Mahalanobis distance ($D_m$) between the anticipated and detected center, similar to the DeepSORT method \cite{Wojke2017simple}. We then combine the Mahalanobis distance with the cosine distance computed on \gls{reid} features into a singular distance measure ($D$) using the formula $D = \lambda D_r + (1 - \lambda)D_m$, where $\lambda$ is a pre-determined weighting parameter set to $0.98$ in our experiments. The Mahalanobis distance is manually set to infinity if it exceeds a certain threshold, which aligns with the JDE protocol \cite{wang2020towards} and prevents the tracking of trajectories exhibiting implausible motion. We then use the Hungarian algorithm with a matching threshold $\tau_1 = 0.4$ to conclude the first matching stage.

The second stage involves attempting to match undetected boxes and tracklets based on the center distance of their respective boxes, with an increased matching threshold $\tau_2 = \qty{2.5}{\meter}$. We continually update the appearance features of the tracklets at each timestep to account for potential variations in appearance. Any unmatched centers are classified as new tracks, and unmatched tracklets are retained for $10$ timesteps to facilitate recognition if they reemerge later.

%% file: tab/detection.tex
\begin{table*}[th]
\centering
\setlength{\tabcolsep}{4pt}
\begin{tabular}{r|rcccccccc}
\toprule
\multicolumn{2}{c}{} & \multicolumn{4}{c}{Wildtrack} & \multicolumn{4}{c}{MultiviewX} \\\cmidrule(lr){3-6}\cmidrule(lr){7-10}
\multicolumn{2}{c}{}& MODA & MODP  & Precision  & Recall & MODA  & MODP  & Precision  & Recall \\\midrule
\multirow{4}{*}{\rotatebox[origin=c]{90}{Two-Stage}}&RCNN \& Cluster \cite{xu2016multi} & 11.3  & 18.4  & 68    & 43     & 18.7  & 46.4  & 63.5  & 43.9   \\ 
&DeepMCD \cite{chavdarova2017deep}     & 67.8  & \underline{64.2}  & 85    & \underline{82}     & 70.0  & \underline{73.0}  & 85.7  & \underline{83.3}   \\ 
&Deep-Occlusion \cite{baque2017deep}   & \underline{74.1}  & 53.8  & \underline{95}    & 80     & \underline{75.2}  & 54.7  & \underline{97.8}  & 80.2   \\ 
&MVTT \cite{lee2023multi}              & \textbf{94.1}  & \textbf{81.3}  & \textbf{97.6} & \textbf{96.5}   & \textbf{95.0} & \textbf{92.8} & \textbf{99.4}  & \textbf{95.6} \\
\midrule
\multirow{5}{*}{\rotatebox[origin=c]{90}{One-Stage}}&MVDet \cite{hou2020multiview}          & 88.2  & 75.7  & 94.7  & 93.6   & 83.9  & 79.6  & 96.8  & 86.7   \\
&SHOT \cite{song2021stacked}           & 90.2  & 76.5  & \underline{96.1}  & 94.0   & 88.3  & 82.0  & 96.6  & 91.5 \\
&3DROM$^\dagger$ \cite{qiu20223d}      & \underline{91.2}  & 76.9  & 95.9  & \underline{95.3}   & 90.0  & 83.7  & 97.5  & 92.4 \\
&MVDeTr  \cite{hou2021multiview}       & \textbf{91.5}  & \textbf{82.1} & \textbf{97.4} & 94.0   & \underline{93.7}  & \textbf{91.3}  & \textbf{99.5}  & \underline{94.2}   \\
&\textbf{\sname}                       & \underline{91.2} & \underline{81.8} & 94.9 & \textbf{96.3} & \textbf{94.2} & \underline{90.1} & \underline{98.6} & \textbf{95.7} \\
\bottomrule
\end{tabular}
\caption{Evaluation of the detection performance with the state-of-the-art methods on the Wildtrack and MultiviewX datasets.  $^\dagger$ 3DROM results are without additional data augmentations.}
\label{tab:detection-sota}
\end{table*}

%% file: tab/tracking.tex
\begin{table}[t]
\setlength{\tabcolsep}{2.75pt}
\centering
\resizebox{\linewidth}{!}{%
\begin{tabular}{rcccccc}
\toprule
& \multicolumn{5}{c}{Wildtrack}\\\cmidrule(lr){2-6}
& IDF1$\uparrow$ & MOTA$\uparrow$ & MOTP$\uparrow$ & MT$\uparrow$ & ML$\downarrow$ \\
\midrule
\small{KSP-DO~\cite{chavdarova2018wildtrack}}        & 73.2 & 69.6 & 61.5 & 28.7 & 25.1\\ 
\small{KSP-DO-ptrack~\cite{chavdarova2018wildtrack}} & 78.4 & 72.2 & 60.3 & 42.1 & 14.6\\ 
\small{GLMB-YOLOv3~\cite{ong2020bayesian}}           & 74.3 & 69.7 & 73.2 & 79.5 & 21.6\\ 
\small{GLMB-DO     \cite{ong2020bayesian}}           & 72.5 & 70.1 & 63.1 & \textbf{93.6} & 22.8\\ 
\small{DMCT  \cite{you2020real}}                     & 77.8 & 72.8 & 79.1 & 61.0 & \textbf{4.9}\\ 
\small{DMCT Stack  \cite{you2020real}}               & 81.9 & 74.6 & 78.9 & 65.9 & \textbf{4.9}\\
\small{ReST$^\dagger$  \cite{cheng2023rest}}         & \underline{86.7} & \underline{84.9} & \underline{84.1} & \underline{87.8} & \textbf{4.9}\\\midrule
\small{\textbf{\sname}}                              & \textbf{92.3} & \textbf{89.5} & \textbf{86.6} & 78.0 & \textbf{4.9}\\
\midrule
& \multicolumn{5}{c}{MultiviewX}\\\cmidrule(lr){2-6}
& IDF1$\uparrow$ & MOTA$\uparrow$ & MOTP$\uparrow$ & MT$\uparrow$   & ML$\downarrow$ \\
\midrule
\small{\textbf{\sname}} & \textbf{82.4} & \textbf{88.4} & \textbf{86.2} & \textbf{82.9} & \textbf{1.3} \\
\bottomrule
\end{tabular}}
\caption{Evaluation of tracking results on the Wildtrack and MultiviewX.
$^\dagger$ ReST originally reported the tracking metrics on view-based tracking instead of tracking in the projected view. The results shown are re-computed by us.}
\label{tab:track-sota}
\end{table}

%% file: sec/4_experiments.tex
\section{Experiments}\label{sec:experiments}

\subsection{Dataset \& Metrics}

\nparagraph{Wildtrack Dataset} Wildtrack~\cite{chavdarova2018wildtrack} is a real-world dataset captured using seven synchronized and calibrated cameras with an overlapping field-of-view of an area of \qty{12}{\metre} $\times$ \qty{36}{\metre}. The movement of the pedestrians is in a public environment and unscripted. Annotations are provided on the ground plane quantized into a $480 \times 1440$ grid, resulting in grid cells of $\qty{2.5}{\cm} \times \qty{2.5}{\cm}$.
The average number of pedestrians per frame is 20, and 3.74 cameras cover each location. Each camera image is recorded at a resolution of $1080 \times 1920$ pixels with a frame rate of \qty{2}{fps}, covering a total of \qty{35}{\min}.

\nparagraph{MultiviewX Dataset} MultiviewX~\cite{hou2020multiview} is a synthetic dataset generated in a game engine and is built to be a synthetic copy of the Wildtrack dataset.
MultiviewX contains views generated by 6 virtual cameras with overlapping field-of-view. The captured area is with \qty{16}{\metre} $\times$ \qty{25}{\metre} slightly smaller than the area of the Wildtrack dataset. For annotation, the ground plane is quantized into a grid of size $640 \times 1000$, where each grid represents the same $\qty{2.5}{\cm} \times \qty{2.5}{\cm}$ squares. The average number of pedestrians per frame is 40, while 4.41 cameras cover each location. The camera resolution ($1080 \times 1920$), frame rate (\qty{2}{fps}), and the length (\qty{400}{frames}) are equal to Wildtrack.

\nparagraph{Detection Metrics}
Unlike monocular-view detection systems, which evaluate the predicted bounding boxes, multi-view detection systems assess the projected ground plane occupancy map. Thus, the comparison to the ground truth is not calculated with the \gls{iou} but with the Euclidean distance as proposed in \cite{chavdarova2018wildtrack}. Detection is classified as true positive if it is within a distance $r = \qty{0.5}{\metre}$, which roughly corresponds to the radius of a human body. Following previous works \cite{chavdarova2018wildtrack, hou2020multiview}, we use \gls{moda} as the primary performance indicator, as it accounts for the normalized missed detections and false positives. Additionally, we report the \gls{modp}, Precision, and Recall.

\nparagraph{Tracking Metrics}
For tracking the metrics are also calculated in the ground plane. We report the common \gls{mot} metrics \cite{bernardin2008evaluating} and identity-aware metrics \cite{ristani2016performance}, the threshold for a positive assignment is set to $r = \qty{1}{\metre}$ to normalize the \gls{motp}. The primary metrics under consideration are \gls{mota} and IDF1.
\gls{mota} takes missed detections, false detections, and identity switches into account. IDF1 measures missed detections, false positives, and identity switches.
Additionally, we also report Mostly Tracked (MT) and Mostly Lost (ML). These are reported as a percentage of the total count of unique pedestrians present in the test set.

\input{tab/model_ablation}
\input{tab/backbone_ablation}

\begin{figure*}[t]
\vskip0.5em
  \centering
  \begin{subfigure}[b]{\textwidth}
    \centering
    \begin{minipage}{0.49\textwidth}
      \includegraphics[width=\textwidth]{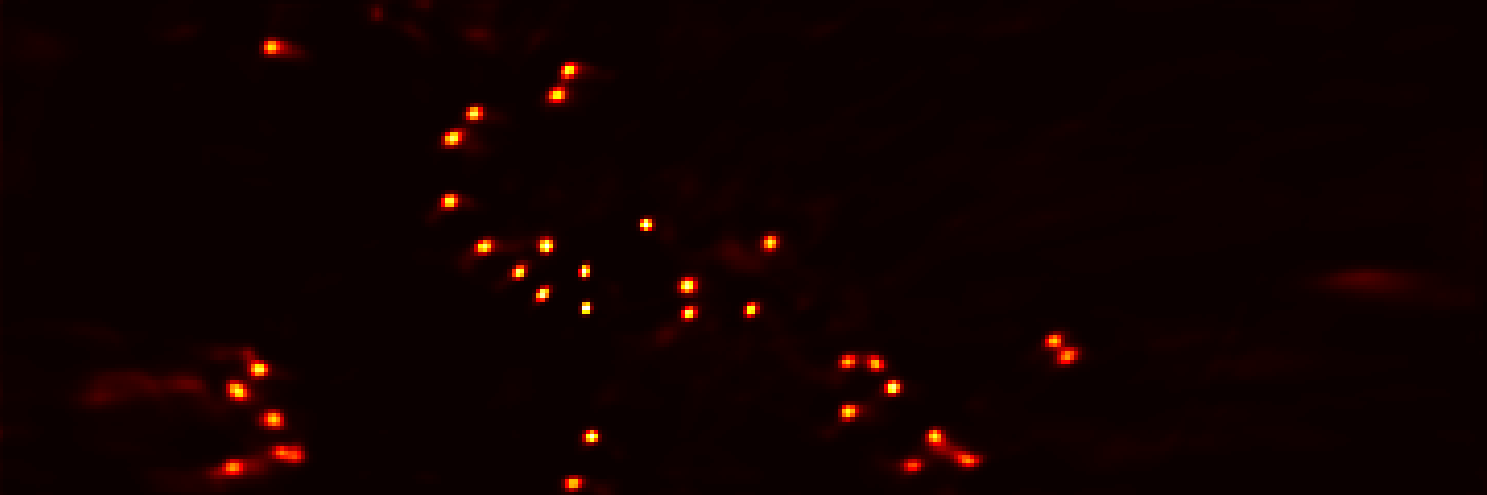}
    \end{minipage}\hfill
    \begin{minipage}{0.49\textwidth}
      \includegraphics[width=\textwidth]{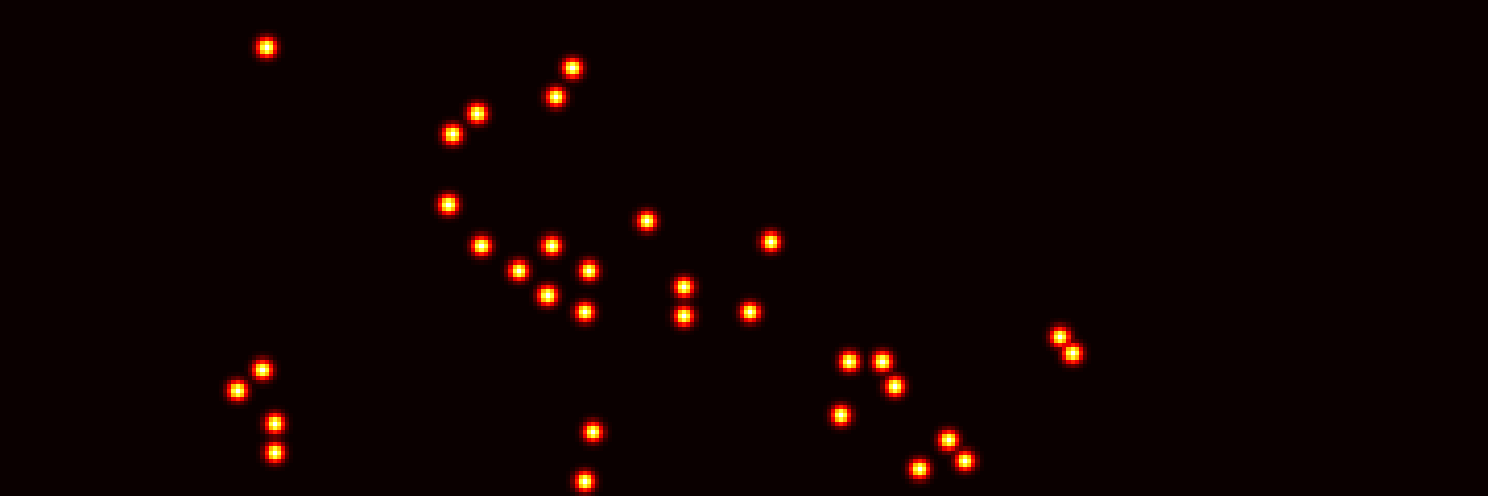}
    \end{minipage}\vspace{0.5em}
    \caption{Comparison of the detection result on Wildtrack as a heat map. Each point in the ground truth represents a pedestrian in the \gls{bev}.}
    \label{fig:quality-det}
  \end{subfigure}\vspace{1em}
  \begin{subfigure}[b]{\textwidth}
    \centering
    \begin{minipage}{0.49\textwidth}
      \includegraphics[width=\textwidth]{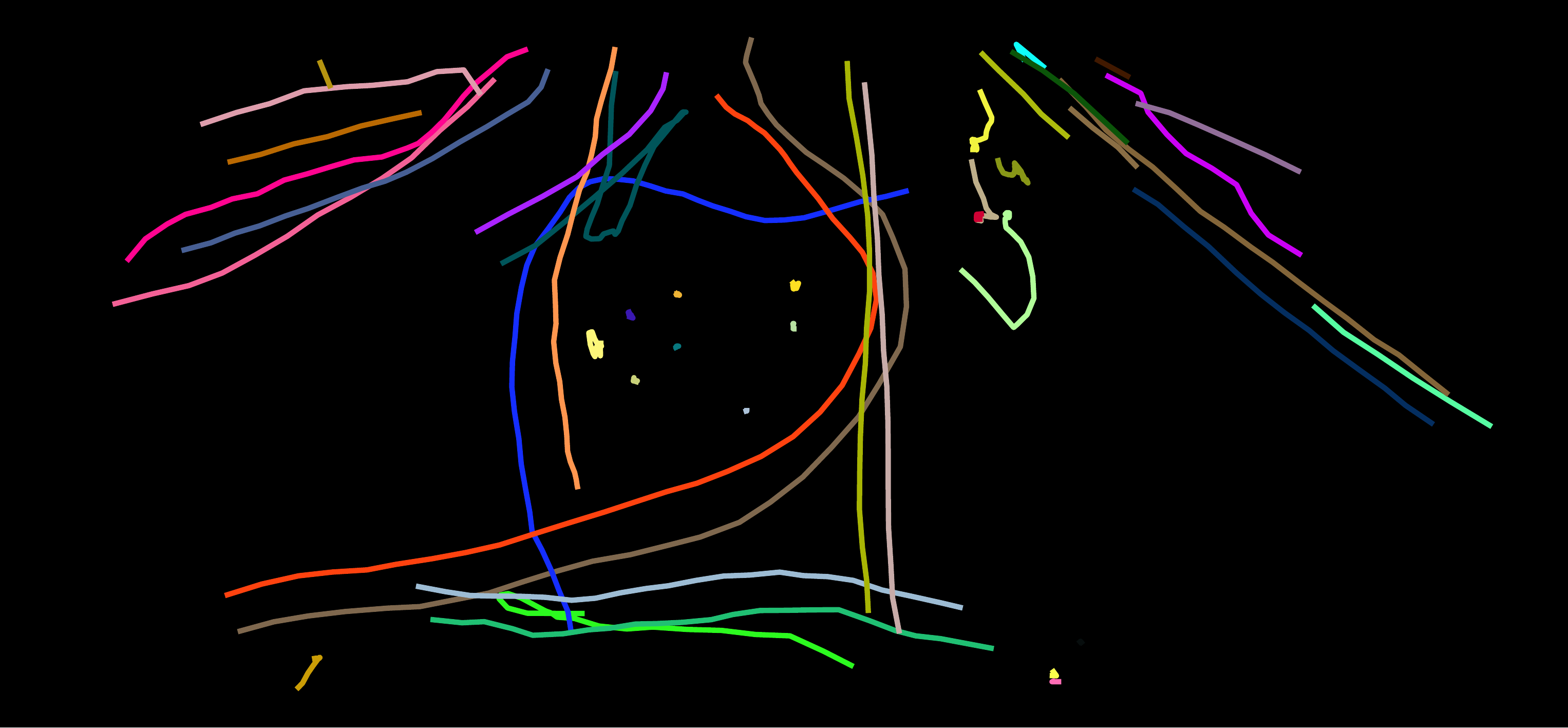}
    \end{minipage}\hfill
    \begin{minipage}{0.49\textwidth}
      \includegraphics[width=\textwidth]{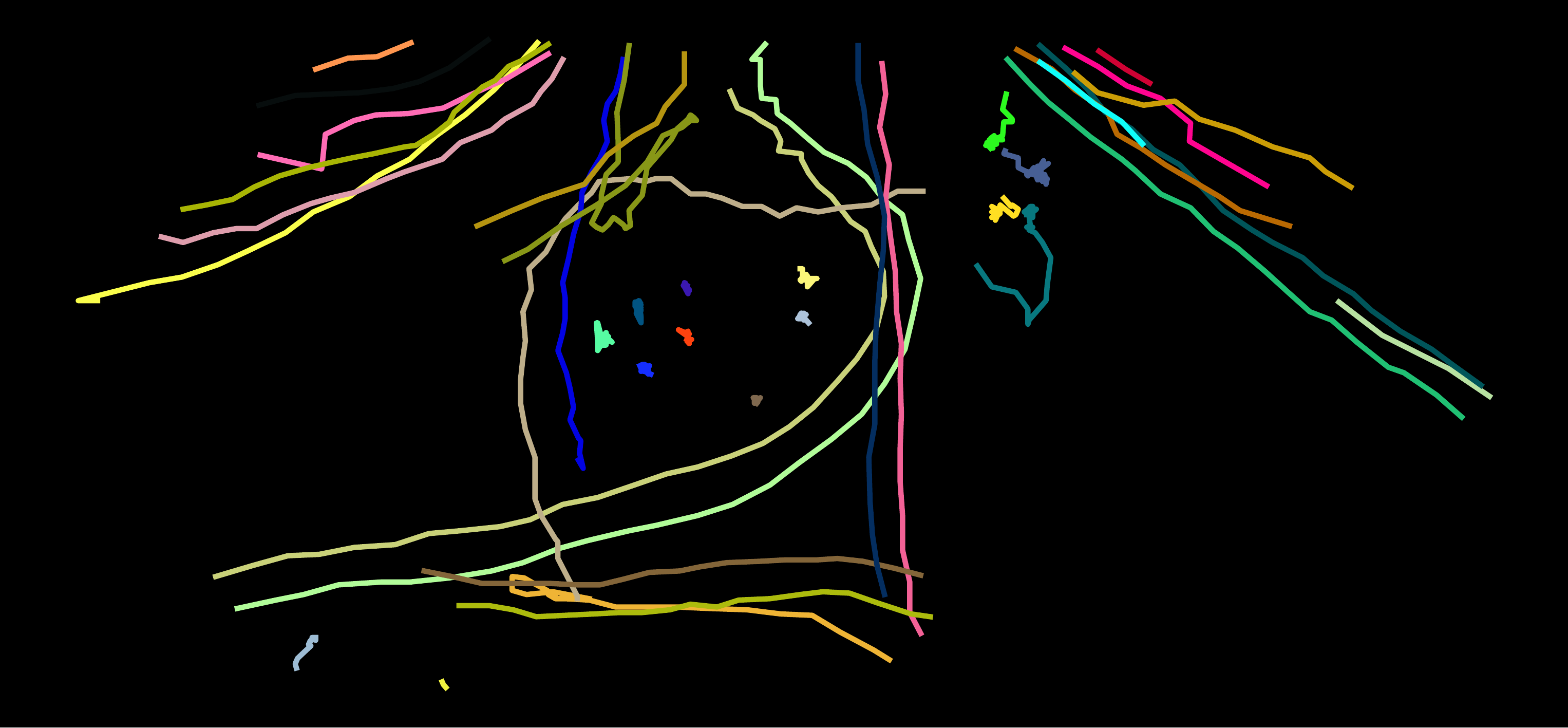}
    \end{minipage}\vspace{0.5em}
    \caption{Comparisons of the tracking results on Wildtrack. Each colored line represents a path taken by one tracked pedestrian as seen from the \gls{bev}.}
    \label{fig:quality-track}
  \end{subfigure}
  \caption{Qualitative detection (a) and tracking (b) results of our approach (left) compared to the ground truth (right).}
  \label{fig:quality}
\end{figure*}

\subsection{Implementation Details}
The input size of the images is $720 \times 1280$ pixels. For augmentation at train time, we follow \cite{harley2022simple, hou2021multiview}: we apply random resizing and cropping on the RGB input, in a scale range of $[0.8, 1.2]$ and adapt the camera intrinsic $K$ accordingly. Additionally, we add some noise to the translation vector $\bm{t}$ of the camera extrinsic to avoid overfitting the decoder. 
We train the detector using an Adam optimizer with a one-cycle learning rate scheduler with a maximum learning rate of $10^{-3}$. We train for $50$ epochs and depending on the size of the encoder, with a batch size of $1-2$ but accumulate gradients over multiple batches before updating the weights to have a effective batch size of $16$. The encoder and decoder network are initialized with pre-trained weights on \textit{ImageNet-1K}.
All experiments are conducted using one RTX 3090 GPU.

\subsection{Main Results}

\nparagraph{Detection} In the tracking-by-detection paradigm, good detections are the basis for good tracking results. While our method does not focus on improving the detection, we still need to be close to the state-of-the-art to achieve competitive results. \cref{tab:detection-sota} compares our methods detection performance to previous methods. We first compare our results to the baseline: MVDet \cite{hou2020multiview}, as we base our approach on it. The results show that our decoder architecture and augmentation changes improved MVDet. Other detection-focused methods \cite{hou2021multiview, lee2023multi, song2021stacked} also extend MVDet and achieve comparative results on Wildtrack, but our approach has competitive results on MultiviewX. The current state-of-the-art MVTT \cite{lee2023multi} is a two-stage detection approach and could still be added to our single-state approach to further improve the results.

\nparagraph{Tracking} In \cref{tab:track-sota} we compare our method to state-of-the-art approaches. Our approach outperforms all current approaches by a big margin. Compared to the current best-performing method ReST \cite{cheng2023rest}, we improve the IDF1 by $5.6$, and the MOTA by $4.6$ percent points. All other methods \cite{chavdarova2018wildtrack, nguyen2022lmgp, ong2020bayesian, you2020real, cheng2023rest} start from 2D detections, and to compare tracking methods, you also need to take the detection quality into account. For most of these approaches, we cannot directly compare the detection quality, but ReST \cite{cheng2023rest} uses detection from MVDeTr \cite{hou2021multiview}, which performs very close to our detector (\cf \cref{tab:detection-sota}). ReST and \sname follow a similar approach: associate spatially on the ground plane, then associate temporally. However, ReST only projects detections in 2D to the ground plane and associates them with a graph solver. In contrast, we project the complete input image feature space to the ground plane and associate it with the decoder. Our results show that with similar detection quality, our approach outperforms ReST, which shows the advantage of our early-fusion approach compared to graph-based late-fusion.

\subsection{Ablations Studies}
\nparagraph{Influence of Method Components}
Next, we ablate each component introduced in our method compared to the baseline as shown in \cref{tab:model-ablation}. The baseline consists of MVDet~\cite{hou2020multiview} with minimal additions to perform tracking, namely a ReID head added to the \gls{bev}-space. The baseline results suffer from strong overfitting and the added augmentation in the next step mostly aleviates this. We augment the input images with scaling and cropping to avoid overfitting in the encoder and add transitive noise to the projection to the ground plane to avoid overfitting in the prediction heads. With better augmentation, we introduced our larger decoder based on a ResNet-18 with a feature pyramid. These additions gave us one of the most robust detection results, but as tracking is our primary focus, we added the view center and \gls{reid} loss. These two losses are applied to the image features and should help guide these features. While the 2D center detection alone decreased our detection performance, using it with the \gls{reid} loss gave us the final SOTA results.

\nparagraph{Influence of the Encoder Network} The encoder extracts the features of the RGB image that are projected to the ground plane. While all of the similar detection-based approaches \cite{hou2020multiview, hou2021multiview, lee2023multi, qiu20223d} use a ResNet-18 encoder, these approaches do not need to encode the identity feature. Thus, we try our approach with larger encoders and transformer-based encoders, see \cref{tab:backbone-ablation}. The ablation shows that ResNet-18 has the best performance. ResNet-50 may be slightly better in the detection and tracking performance, but the smaller ResNet-18 outperforms it in the main metrics MODA and MOTA and has competitive scores for IDF1. We thus use ResNet-18 to report all other results.


\subsection{Qualitative Results}
In \cref{fig:quality}, we plot the output of our model on the test set of Wildtrack. In \cref{fig:quality-det}, we compare the prediction of the \gls{pom} to the ground truth as a heatmap at a single timestep. Each point in the ground truth represents a pedestrian in the \gls{bev}. The scene's center is crowded with pedestrians, and the prediction in this high-overlap area is almost perfect. The further the pedestrians are on a side, the less accurate the prediction is. This inaccuracy could be due to the increased distortion further from the cameras and less overlap of the views in the border regions. 

\cref{fig:quality-track} shows similar results for tracking. We show all tracks on the ground plane of the full test set of Wildtrack, where each color and line represents the path one identity takes. The tracks in the center of the scene are predicted almost perfectly. However, the top-left and top-right tracks are segmented or switched tracks. This inaccuracy could be due to less accurate or missing detections and identity features outside the ground plane.

%% file: tab/model_ablation.tex
\begin{table}[t]
\setlength{\tabcolsep}{2.75pt}
\centering
\begin{tabular}{lcccccc}
\toprule
& \multicolumn{2}{c}{Detection} & \multicolumn{3}{c}{Tracking} \\\cmidrule(lr){2-3}\cmidrule(lr){4-6}
& MODA & MODP & IDF1 & MOTA & MOTP  \\
\midrule
\small{Baseline} &  77.8 & 78.9 & 71.3 & 72.6 & 80.9\\  
\midrule
$+$ \small{Augmentation}         & 89.5 & 81.7 & 84.5 & 87.4 & 83.0\\
$+$ \small{Decoder}              & \textbf{91.3} & \textbf{82.2} & \underline{91.1} & \underline{89.1} & \textbf{86.9}\\
$+$ \small{View Center Loss}     & 91.0 & \underline{82.1} & 90.0 & \underline{89.1} & 84.0\\
$+$ \small{View \gls{reid} Loss} &\underline{91.2} & 81.8 & \textbf{92.3} & \textbf{89.5} & \underline{86.6}\\
\bottomrule
\end{tabular}
\caption{Ablation of the components introduced by our approach compared to the baseline method.}
\label{tab:model-ablation}
\end{table}

%% file: tab/backbone_ablation.tex
\begin{table}[t!]
\setlength{\tabcolsep}{2.75pt}
\centering
\begin{tabular}{rcccccc}
\toprule
& \multicolumn{2}{c}{Detection} & \multicolumn{3}{c}{Tracking} \\\cmidrule(lr){2-3}\cmidrule(lr){4-6}
& MODA & MODP & IDF1 & MOTA & MOTP  \\
\midrule
ResNet-18 & \textbf{91.2} & \underline{81.8} & \underline{92.3} & \textbf{89.5} & \underline{86.6} \\
ResNet-50 & \underline{89.6} & \textbf{82.3} & \textbf{92.6} & \underline{88.8} & 86.5 \\
Swin-T   & 89.5 & 81.3 & 92.0 & 87.3 & \textbf{87.9} \\
\bottomrule
\end{tabular}
\caption{Ablation of different encoders on detection and tracking results of the Wildtrack dataset.}
\label{tab:backbone-ablation}
\end{table}

%% file: sec/5_conclusion.tex
\section{Discussion}

\nparagraph{Limitations} The first limitation of our approach is the requirement for high-quality 3D annotations and camera calibrations. While this is easy to archive with synthetic data, it is costly for real-world data. Therefore, we could not evaluate some older datasets (CAMPUS~\cite{xu2016multi}, PETS09~\cite{ferryman2009pets2009}), where most of the late-fusion models can work with only 2D annotations and ground plane homography.
Furthermore, our approach requires synchronized cameras. Since we lift all cameras to the same 3D space, the time differences should be minimal so that moving objects do not project to different locations in 3D. Late-fusion \gls{mtmc} tracking methods can account for more drift in the temporal domain. Our approach also has higher hardware requirements. While late-fusion approaches may process each camera decentralized and fuse the information centrally, our approach processes all camera images simultaneously. It thus requires more memory and computational resources on a single machine.

\nparagraph{Ethical Impact}%
Tracking methods always have the risk of being used for illegal surveillance. Methods that focus on pedestrian tracking must face this criticism, especially. The dataset Wildtrack~\cite{chavdarova2018wildtrack} has been criticized~\cite{Exposing.ai} for the missing consent of the recorded persons. Unfortunately, a good comparison to the state-of-the-art is only possible with this dataset, even though a synthetic replication with MultiviewX~\cite{hou2020multiview} is now available. We are the first to evaluate tracking on this synthetic dataset to lower the ethical implications of tracking.

\nparagraph{Future Work}
For the detection part, the biggest challenge of our and other current approaches \cite{hou2021multiview, lee2023multi, song2021stacked} is the distortion caused by the projective transformation. Other methods that lift from 2D to 3D space could be explored for multi-view detection, i.e., Simple-BEV \cite{harley2022simple}, Lift-Splat-Shot \cite{philion2020lift}, or BEVFormer \cite{li2022bevformer}.
Most current approaches only use the current frame for detection. Using more temporal frames could improve the detection performance \cite{li2022bevformer}. Using more temporal context could also improve the tracking quality, and approaches like CenterTrack~\cite{zhou2020tracking} could be used to track via motion cues.
The pedestrian datasets used in this work have about $400$ timestamps, which is relatively small by modern computer vision standards, and the detection and tracking accuracy is saturating. The need for larger datasets is apparent, and datasets with similar \gls{mtmc} problems for traffic surveillance \cite{tang2019cityflow, synthehicle2023herzog} could bridge this gap. 

\section{Conclusion}\label{sec:conclusion}
This paper shows that the \sname catches the worm through multi-view highly accurate tracking. Early-fusion of all views and tracking in the bird's eye view considerably improves \gls{mtmc} tracking. We adapt one-shot tracking to multi-view tracking to propose an online, anchor-free tracker. We propose ways to efficiently train \gls{reid} features in \gls{bev} and ablate each of our tracking improvements.

We expect \sname to inspire feature work in early-fusion multi-view tracking and believe that \sname, together with our suggested future work, makes significant progress towards tackling multi-view tracking problems.